\documentclass[11pt,a4paper]{article}

\usepackage[round,authoryear]{natbib}
\usepackage{times}

\usepackage[margin=26mm,top=24mm,bottom=25mm]{geometry}

\usepackage[%
pdfstartview=FitH,%
breaklinks=true,%
bookmarks=false,%
colorlinks=true,%
linkcolor= blue,
anchorcolor=blue,%
citecolor=blue,
filecolor=blue,%
menucolor=blue,%
urlcolor=blue%
]{hyperref}
\usepackage{xcolor}

\definecolor{blue}{HTML}{0077BB}
\definecolor{cyan}{HTML}{33BBEE}
\definecolor{green}{HTML}{009988}
\definecolor{orange}{HTML}{EE7733}
\definecolor{red}{HTML}{CC3311}
\definecolor{magenta}{HTML}{EE3377}
\definecolor{grey}{HTML}{BBBBBB}
\definecolor{pink}{HTML}{FF007F}

\usepackage{graphicx}
\usepackage{amsmath}
\usepackage{amssymb}
\usepackage{comment}
\usepackage{caption}
\usepackage{subcaption}
\usepackage{algorithm}
\usepackage{algorithmic}
\usepackage{newfloat}
\usepackage{listings}
\DeclareCaptionStyle{ruled}{labelfont=normalfont,labelsep=colon,strut=off}
\floatstyle{ruled}
\newfloat{listing}{tb}{lst}{}
\floatname{listing}{Listing}
\usepackage{booktabs}
\usepackage{hyperref}
\usepackage{cleveref}
\usepackage{tikz}
\usetikzlibrary{calc}
\usepackage{pifont}
\usepackage{float}
\usepackage{enumitem}

\newcommand{\radaraxis}[1]{%
    \foreach \frac in {0.2,0.4,0.6,0.8,1.0}{
        \draw[gray!40] (90:\frac*#1) -- (162:\frac*#1) -- (234:\frac*#1)
                        -- (306:\frac*#1) -- (18:\frac*#1) -- cycle;
    }
    \foreach \ang in {90,162,234,306,18}{
        \draw[gray!60] (0,0) -- (\ang:#1);
    }
    \node[align=center, font=\scriptsize] at (90:#1+0.75)  {Cond.\\Validity};
    \node[align=center, font=\scriptsize] at (162:#1+0.9)  {FID\\Target};
    \node[align=center, font=\scriptsize] at (234:#1+0.9)  {FID\\Source};
    \node[align=center, font=\scriptsize] at (306:#1+0.75) {FLOPs/\\sample};
    \node[align=center, font=\scriptsize] at (18:#1+0.9)   {Model\\Size};
}

\newcommand{\xbf}{\mathbf{x}}

\newcommand{\Ebb}{\mathbb{E}}
\newcommand{\R}{\Rbb}
\newcommand{\Mbb}{\mathbb{M}}
\newcommand{\M}{\Mbb}
\newcommand{\ie}{\emph{i.e.}}
\newcommand{\eg}{\emph{e.g.}}
\newcommand{\Ncal}{\mathcal{N}}
\newcommand{\Ucal}{\mathcal{U}}
\newcommand{\Lcal}{\mathcal{L}}

\newcommand{\zerobf}{\mathbf{0}}
\newcommand{\Ibf}{\mathbf{I}}
\DeclareMathOperator*{\EE}{\Ebb}
\DeclareMathOperator*{\argmin}{argmin}

\title{CounterFlow: Target-Conditioned\\ Flow Matching for Visual Counterfactuals}

\author{Paul Sevellec$^{1,3,}$\thanks{Corresponding Author. Email: prenom.nom@inria.fr} 
\quad Elisa Fromont$^{1}$
\quad Romaric Gaudel$^{1}$ \\
Laurence Roze$^{2}$ 
\quad Paul Viallard$^{1}$ 
\quad Laetitia Houeix$^{3}$ \\
$^{1}$ Univ Rennes, CNRS, Inria, IRISA - UMR 6074, F35000 Rennes, France \\
$^{2}$ Univ Rennes, INSA Rennes, CNRS, Inria, IRISA - UMR 6074, F35000 Rennes, France \\
$^{3}$ Stellantis Inc. 78300 Poissy, France
}
\date{}

\begin{document}

\maketitle

\begin{abstract}
In eXplainable AI, counterfactual explanations have gained significant traction by offering users direct, intuitive, and actionable insights into model decisions. We introduce CounterFlow, a novel framework based on Conditional Flow Matching that generates visual counterfactual explanations by directly steering a source image toward a user-specified target class. By leveraging target-conditioned vector fields, CounterFlow identifies highly realistic semantic modifications while requiring substantially lower computational overhead than traditional GAN- or diffusion-based architectures. Empirical evaluations on standard benchmarks demonstrate that CounterFlow achieves state-of-the-art performance in both visual realism and target class switching. By working directly in the raw pixel space, CounterFlow offers a more practical solution for real-world deployment.
\end{abstract}

\section{Introduction}

\begin{figure}[t]
    \centering
    \begin{subfigure}[t]{0.95\linewidth}
        \centering
        \footnotesize
        \makebox[0.2\linewidth][c]{$7\to9$}%
        \makebox[0.2\linewidth][c]{$5\to0$}%
        \makebox[0.2\linewidth][c]{$4\to5$}%
        \makebox[0.2\linewidth][c]{$3\to2$}%
        \makebox[0.2\linewidth][c]{$2\to3$}\\[2pt]
        \includegraphics[width=\linewidth]{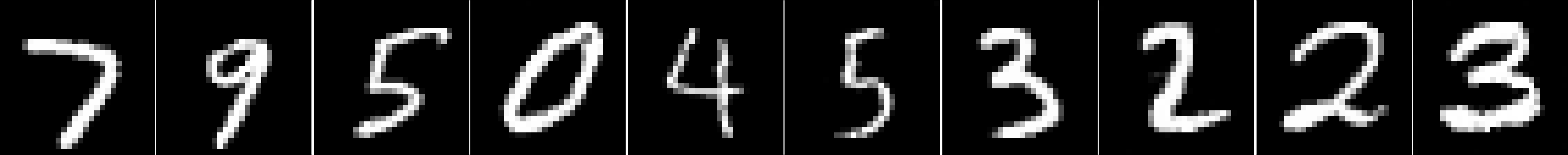}
        \subcaption{MNIST}
        \label{fig:mnist_cifar:mnist}
    \end{subfigure}
    \vspace{6pt}

    \begin{subfigure}[t]{0.95\linewidth}
        \centering
        \footnotesize
        \makebox[0.2\linewidth][c]{dog$\to$bird}%
        \makebox[0.2\linewidth][c]{horse$\to$truck}%
        \makebox[0.2\linewidth][c]{dog$\to$horse}%
        \makebox[0.2\linewidth][c]{bird$\to$deer}%
        \makebox[0.2\linewidth][c]{airplane$\to$auto}\\[2pt]
        \includegraphics[width=\linewidth]{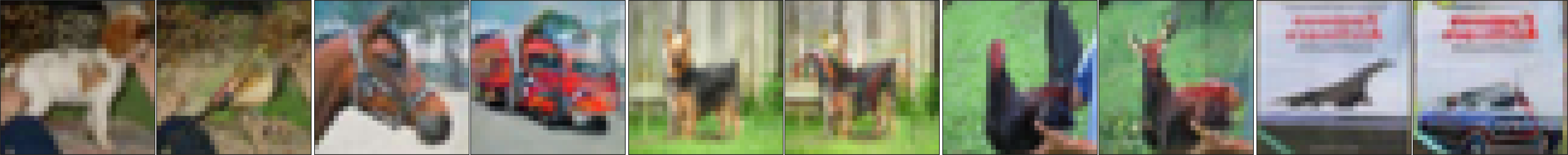}
        \subcaption{CIFAR10}
        \label{fig:mnist_cifar:cifar}
    \end{subfigure}
    \caption{%
      Qualitative counterfactual explanations generated by \textbf{CounterFlow} on (\subref{fig:mnist_cifar:mnist})~MNIST and (\subref{fig:mnist_cifar:cifar})~CIFAR10.
      We show five examples per dataset as \emph{source}/\emph{counterfactual} pairs (\emph{source} on the left of each pair, the corresponding \emph{counterfactual} generated by CounterFlow on the right), and we indicate the source$\to$target class transition above each pair.
      CounterFlow preserves source-specific style, such as stroke thickness and digit slant on MNIST, or object color and orientation on CIFAR10, modifying only what is needed to reach the target class.
    }
    \label{fig:mnist_cifar_qualitative}
\end{figure}

\textbf{Context.}
On the one hand, in generative modeling, diffusion models currently dominate the landscape for high-quality image, video, and audio synthesis \citep{rombach2022high-resolution,ho2022video,liu2023audioldm}.
However, they are also well-known for their slow training times and their stochastic sampling paths.
To address these limitations, \citet{lipman2023flow,albergo2023building,liu2023flow} introduced Conditional Flow Matching (CFM).
CFM is a generative modeling framework based on Continuous Normalizing Flows (CNFs) that uses standard regression objectives for training.
Moreover, instead of relying on diffusion processes, CFM generates examples by integrating the CNF's ODE along deterministic trajectories.
On the other hand, the generative models' ability to generate complex data distributions and thus to produce realistic examples is particularly appealing for eXplainable Artificial Intelligence \citep[XAI,][]{dwivedi2023explainable}, where one aims to produce counterfactual explanations \citep{wachter2017counterfactual}, \ie, examples \textit{(i)} close to a given input, \textit{(ii)} plausible, and \textit{(iii)} that change the model's prediction.
Unlike adversarial examples that rely on imperceptible noise to fool the classifier, counterfactual examples expose meaningful changes, helping users to trust the model while enabling developers to debug it.
Our work leverages CFM to create counterfactual explanations: rather than generating images from random noise, our approach transports an input sample toward a counterfactual conditioned on a user-defined target class, while jointly optimizing for plausibility and proximity; see \Cref{fig:mnist_cifar_qualitative}.

\textbf{Related Work.}
Prior work on image counterfactual explanations falls into two categories: instance-based and generative approaches.
Instance-based methods \citep[see \eg,][]{goyal2019counterfactual, keane2020good} select examples from the training set using similarity metrics to modify the input sample $\xbf$.
However, they require access to the training set at inference and cannot tailor a minimal, plausible edit to a specific example $\xbf$.
We focus instead on generative approaches, a more recent line of work \citep{khorram2022cycle-consistent, jeanneret2024diffusion, weng2024fast, augustin2022diffusion, cao2025leapfactual} that produces more plausible examples, which we group by generative backbone: GANs, diffusion models, or CFM.
\begin{itemize}[leftmargin=*, itemsep=2pt, topsep=2pt, parsep=0pt]
    
    \item \textbf{GAN.}
C3LT~\citep{khorram2022cycle-consistent} encodes the input (with a VAE encoder) into a latent space, in which it learns two cycle-consistent transformations per pair of classes $(y, y')$ before decoding with the generator of a GAN.
It thus requires one model set per pair, impractical for many classes, and its fidelity is bounded by the VAE's reconstruction quality.

\item \textbf{Diffusion.}
DiME \citep{jeanneret2024diffusion} generates a counterfactual by corrupting the input and denoising it through the backward diffusion process (an inner loop of several steps); at each outer step, the classifier gradient computed on this denoised estimate refines the corrupted instance, and the process repeats until it converges to a valid counterfactual.
FastDiME \citep{weng2024fast} instead computes this gradient directly from the model's one-step denoised estimate, bypassing the inner loop entirely.
ACE \citep{jeanneret2023adversarial} takes a different route: it builds an adversarial attack against the classifier and lets the diffusion model's own denoising process project this attack back onto the manifold of realistic images, turning an imperceptible perturbation into a semantically meaningful one.
DVCE \citep{augustin2022diffusion} starts the reverse diffusion from a partially noised version of the input, rather than pure noise, and steers the process by projecting the gradient of an adversarially robust classifier onto a cone around the target classifier's own gradient direction whenever the two disagree, avoiding non-semantic changes.
More recently, MaskDiME \citep{guo2026maskdime} improves on DiME's efficiency and spatial precision with a training-free, adaptive dual-mask mechanism guided by classifier gradients that restricts updates to a localized region rather than the whole image.
Regardless of the specific guidance scheme, diffusion-guidance methods must reapply their guidance at every one of many denoising steps, making inference slow: DiME's inner diffusion loop scales quadratically with the number of diffusion steps, and even FastDiME's one-step approximation, despite yielding a roughly 20$\times$ empirical speedup, only reduces this to a linear scaling and remains markedly slower than a single forward pass. ACE and DVCE compound this same per-step cost further, respectively adding an adversarial attack and a robust-classifier gradient computation at every step.

\item \textbf{CFM.}
LeapFactual~\citep{cao2025leapfactual} builds on Conditional Flow Matching (CFM) to generate counterfactuals. It learns a class-conditional generative flow from a Gaussian prior to the latent space of a pretrained VAE, and repurposes this flow for counterfactual generation through a \emph{lift-and-land} procedure. Given an input, its VAE encoding is first \emph{lifted} toward the prior under the input class by solving the learned ODE backward, and then \emph{landed} back into the latent space under the target class by solving the ODE forward. This ``leap'' must be repeated a dataset-dependent number of times, with each leap requiring a VAE decoding and a classifier query.

Our work also builds on CFM, but uses it in a fundamentally different way. We directly learn a transport from an input image to its counterfactual. This avoids (i) the need for multiple costly leaps, (ii) the intermediate traversal through a prior space, and (iii) VAE encoding and decoding, whose reconstruction fidelity can limit the quality of the generated counterfactuals.

\end{itemize}

\textbf{Our contributions.}
To address the drawbacks of the previous methods, our contributions are as follows:

\begin{itemize}[leftmargin=*, itemsep=2pt, topsep=2pt, parsep=0pt]
    \item We introduce CounterFlow, the first CFM framework that directly conditions the vector field on both the source and target class, enabling a single model to transport examples between any pair of classes in their original pixel representation.
    Moreover, CounterFlow generates a counterfactual by integrating a single (learned) ODE from source to target, with no additional VAE decoding or iterative optimization at inference.

    \item To enable the learning of the transport map, we introduce two complementary contributions: a latent-space nearest-neighbor pairing strategy and an auxiliary classification loss. The pairing algorithm leverages the embedding space of a pretrained VAE to associate each source example with a semantically similar target example, thereby reducing the transport distance that the flow must learn. The auxiliary classification loss further guides the flow toward transformations that induce the desired change in the classifier's prediction.

    \item We provide an empirical evaluation covering conditional validity, plausibility, and computational efficiency, a combination rarely reported together in prior work, demonstrating that CounterFlow matches or improves on state-of-the-art generation quality at a fraction of the computational cost of GAN- or diffusion-based alternatives.
\end{itemize}

\section{Background on Counterfactual Explanations}

Let $X=\R^d$ be the input space and $Y=\{1,\dots,K\}$ the label space. We denote by $\M(H)$ the set of probability distributions on a space $H$.
We assume that there exists a true data distribution $q\in \M(X)$, which is unknown and from which we only have access to a learning sample.
We denote by $\hat{q}$ the empirical distribution over a learning sample $S=\{\xbf^{(i)}\}_{i=1}^m \subset X$, with $S\sim q^m$, so that $\hat{q}=\Ucal(S)$.\footnote{$q^m$ denotes the joint distribution on a sample of size $m$, while $\Ucal$ is a uniform distribution on a given set.}
We also denote by $q_{y}$ the conditional distribution of $q$ given a label $y\in Y$.
Let $f:X\to Y$ be the classification model whose predictions we aim to explain that associates a label $y=f(\xbf)$ to each example $\xbf$.
Our goal is, given $\xbf_0 \sim q_{y_0}$, to generate a counterfactual explanation of $\xbf_0$ (where $y_0=f(\xbf_0)$), denoted by $\xbf_1$; the problem is formulated by \citet{wachter2017counterfactual} (and later reformulated by \citet{artelt2019on}) as
\begin{align*}
\xbf_1 = {\textstyle\argmin_{\xbf'}}\ d(\xbf', \xbf_0) \ \text{s.t.}\  f(\xbf')=y_1 \neq y_0.
\end{align*}
This objective makes $\xbf_1$ an adversarial example: an example close to $\xbf_0$ for the distance $d(\cdot,\cdot)$ that $f$ classifies as $y_1$.

However, this definition can yield implausible examples.
Indeed, according to the true data distribution $q$, sampling $\xbf_1$ may be unlikely, \ie, $q_{y_1}(\xbf_1)$ may be small. Hence, in this paper, we instead adopt the definition of \citet{artelt2020convex}, where we aim to generate a plausible example $\xbf_1$, \ie, one for which $q_{y_1}(\xbf_1)$ is higher than a constant $\varepsilon>0$:
\begin{align}
\xbf_1 = {\textstyle\argmin_{\xbf'}}\ d(\xbf', \xbf_0) \ \text{s.t.}\  f(\xbf') = y_1 \neq y_0\ \text{and}\ q_{y_1}(\xbf')\ge \varepsilon.\label{eq:counterfactual}
\end{align}
\looseness=-1
This remains a difficult problem, as it requires both defining the distance $d$ and solving a generally non-convex optimization problem.
We therefore approximate it: to generate plausible examples $\xbf_1$, we use a conditional flow matching model, introduced in the next section.
Moreover, to indirectly minimize the distance $d$ while respecting the condition $f(\xbf_1)=y_1 \neq y_0$, we further introduce our method: CounterFlow.

\section{Background on Conditional Flow Matching}
\label{background}

\looseness=-1
Generative models aim to approximate $q$ from $\hat{q}$ by starting from a base distribution $p_0 \in \M(X)$ (classically a standard Gaussian $\Ncal(\zerobf, \Ibf_d)$, with $\zerobf$ the zero vector and $\Ibf_d$ the $d{\times}d$-identity matrix) and learning to transport it toward the target distribution $p_1 \in \M(X)$, so that $p_1 = q$.
A Continuous Normalizing Flow \citep[CNF,][]{chen2018neural} is a generative model that achieves this by defining a time-dependent velocity field $v_\theta(\xbf,t)$ parameterized by $\theta$, which generates samples according to
\begin{align}
\frac{d \xbf_t}{dt} = v_\theta(\xbf_t, t), \quad t \in [0,1].\label{eq:ode}
\end{align}
Thanks to this equation, starting from $\xbf_0\sim p_0$, the point $\xbf_1 \sim p_1$ is obtained by solving the integrated version of \Cref{eq:ode}, \ie, we have
\begin{align}
&\xbf_t = \psi_t(\xbf_0) = \xbf_0 + \int_{0}^{t}v_\theta(\xbf_s, s)ds, \ \text{where}\  \xbf_t\sim p_t.\label{eq:int-ode}
\end{align}
In other words, the distribution $p_t$ is defined by the push-forward measure $p_t = (\psi_t)_\# p_0$, and so $\psi_t$ transports samples from the base distribution $p_0$ to the distribution $p_t$.
It defines a probability path $(p_t)_{t \in [0,1]}$ interpolating between a source distribution $p_0$ and a target distribution $p_1$.
Hence, training a CNF amounts to finding $\theta$ such that $p_1$ approximates $q$, given only $\hat{q}$.
However, training CNFs through maximum likelihood can be computationally expensive due to the need to solve ODEs during optimization.
Besides, there exists an infinite number of probability paths, which leads to instability during training \citep[see \eg,][]{lipman2023flow}.
Hence, Flow Matching (FM) reformulates the training of continuous flows as a supervised regression problem on velocity fields. Instead of learning the transport dynamics indirectly through maximum likelihood, FM supervises the velocity field along predefined trajectories connecting the source and target distributions.
The model is then trained to match this target velocity via the objective
\begin{align}
&\Lcal_{\text{FM}}(\theta) =
\EE_{t \sim \Ucal([0,1]), \xbf_t \sim p_t}
\left[
\left\| v_\theta(\xbf_t, t) - v^*(\xbf_t, t) \right\|^2
\right]. \label{eq:fm_loss}
\end{align}
However, $v^*$ depends on the unknown $q$, making $\Lcal_{\text{FM}}(\theta)$ intractable in practice.
Fortunately, \citet[Theorem 2]{lipman2023flow} proved that the loss $\Lcal_{\text{FM}}(\theta)$ has the same gradient as the following loss
\begin{align*}
\Lcal_{\text{CFM}}'(\theta) = \EE_{\substack{t \sim \Ucal([0,1]),\\ z \sim \gamma, \xbf_t \sim p_{t,z}}}
\left[
\left\| v_\theta(\xbf_t, t) - v^*_z(\xbf_t, t) \right\|^2
\right],
\end{align*}
where $z$ is now a conditioning variable drawn from a coupling distribution $\gamma$ that determines which distribution $p_{t,z}$ and target velocity $v^*_z$ govern each sample.
Classically \citep{lipman2023flow,tong2024improving}, $z=(\xbf_0,\xbf_1)\sim\gamma$ is a pair of examples, with
\begin{align}
\xbf_t = (1{-}t)\xbf_0 + t\xbf_1, \quad p_{t,z} = \delta_{\xbf_t}, \quad \text{and} \quad v^*_z(\xbf_t,t) = \tfrac{d\xbf_t}{dt} = \xbf_1-\xbf_0. \label{eq:cond_path}
\end{align}
In other words, the conditional path $p_{t,z} = \delta_{\xbf_t}$ is a Dirac point mass (\ie, a degenerate distribution) supported on $\xbf_t$, and $v^*_z$ is the (unique) velocity field consistent with $p_{t,z}$ through the continuity equation \citep{lipman2023flow}.
Moreover, $p_0$ does not need to be Gaussian nor $\gamma$ to be independent \citep{tong2024improving}: the same construction can transport examples between two data distributions, as we do in CounterFlow (see \Cref{sec:counter_flow} further).

The above objective $\Lcal_{\text{CFM}}'(\theta)$ extends naturally to conditional generation by introducing a context variable $c$ (\eg, a class label), on which both the coupling and the velocity field depend: unlike $z$, which only selects the path for a given pair, $c$ is additionally fed to the network itself so that the learned velocity field can be evaluated for any desired context at inference. The context and the pair are now drawn jointly from the same coupling, $(c,z)\sim\gamma$, and we have
\begin{align}
\label{eq:cfm_loss_cond}
\Lcal_{\text{CFM}}(\theta) = \!\!\!\!
\EE_{\substack{t \sim \Ucal([0,1]),\\ (c,z) \sim \gamma, \, \xbf_t \sim p_{t,z}}}\!\!\!\!
\left[
\left\| v_\theta(\xbf_t, t, c) - v^*_z(\xbf_t, t) \right\|^2
\right].
\end{align}
Note that the path $p_{t,z}$ and target velocity $v^*_z$ remain indexed by $z$ alone, as in the unconditional case, only the learned field $v_\theta(\xbf_t, t, c)$ is additionally conditioned on $c$, which is the mechanism CounterFlow relies on to control the source and target class at inference.

\section{CounterFlow}
\label{sec:counter_flow}

\looseness=-1
This section introduces our method, namely CounterFlow, which learns a velocity field $v_{\theta}$ that generates a counterfactual to approximate \Cref{eq:counterfactual}.
To do so, we define the loss $\Lcal_{\text{CFM}}(\theta)$ derived from the coupling $\gamma$ (and a minibatch version $\gamma^n$) together with the conditional path $p_{t,z}$ of \Cref{eq:cond_path}, where $(c,z)\sim \gamma$; and we end by introducing the training and inference procedures of CounterFlow.

\subsection{Defining the loss}

\textbf{Context $c$ and coupling $\gamma$.}
Since \Cref{eq:counterfactual} transports $\xbf_0$ with $f(\xbf_0)=y_0$ toward a counterfactual $\xbf_1$ with $f(\xbf_1)=y_1$, we define the context as the pair of labels $c=(y_0,y_1)\sim\Ucal(Y^2)$.
Then, we sample two minibatches of size $n$ sharing this context, $B_0\sim\Ucal(S_{y_0})^n$ and $B_1\sim\Ucal(S_{y_1})^n$, where $S_{y} = \{\xbf \in S \;|\; f(\xbf)=y\}$ is the set of examples in $S$ that $f$ classifies as $y$.
Finally, we pair their examples (see below) to form the set $B \in (X^2)^n$ of $n$ pairs $z$.
Overall, \Cref{alg:pair_by_embedding} defines the distribution $\gamma^n$ where $(c,B) \sim \gamma^n$.
Moreover, when we consider only one example $z$ from the minibatch $B$, we say that $(c,z) \sim \gamma$ (as in \Cref{eq:cfm_loss_cond}), which is equivalent to drawing $(c,B)\sim\gamma^n$ and sampling uniformly a single index $I\sim\Ucal(\{1,\dots,n\})$ of the minibatch, \ie, we have $(c,z) \sim \gamma \iff (c,B)\sim\gamma^n$, $z=z^{(I)}$, $I\sim\Ucal(\{1,\dots,n\})$.

\textbf{Latent-space pairing.}
The pairing $B$, created by \Cref{alg:pair_by_embedding}, aims to minimize a distance $d(\xbf_1, \xbf_0)$ where $\xbf_0 \in B_0 = \{\xbf_0^{(i)}\}_{i=1}^{n}$ and $\xbf_1\in B_1 = \{\xbf_1^{(j)}\}_{j=1}^{n}$, which first requires the choice of the distance $d$ itself.
For CounterFlow, we aim for a distance $d$ that measures whether $\xbf_0$ and $\xbf_1$ are semantically related;
hence, we use the latent space of a frozen and pretrained Variational Autoencoder (VAE) and we set
\begin{align}
  d(\xbf_0, \xbf_1) = \| E(\xbf_0) - E(\xbf_1) \|_2, \label{eq:distance}
\end{align}
where $E(\xbf)$ denotes the latent-space embedding of the input $\xbf$.
Then, to obtain the semantically related pairs, for each example $\xbf_0^{(i)} \in B_0$, where $i \in \{1, \dots, n\}$, we associate the closest example $\xbf_1^{(\pi(i))}\in B_1$ according to the distance $d$, where $\pi(i) = {\textstyle\argmin_{j\in\{1,\dots,n\}} d(\xbf_0^{(i)}, \xbf_1^{(j)})}$ is the index in $B_1$ of the nearest neighbor of $\xbf_0^{(i)}$ (see Line~\ref{line:pair-argmin}).

\textbf{Loss $\Lcal_{\mathrm{CFlow}}(\theta)$.}
We are now able to define the loss $\Lcal_{\mathrm{CFlow}}(\theta)$ minimized during the training procedure. This loss is the weighted sum $\Lcal_{\mathrm{CFlow}}(\theta) = \Lcal_{\mathrm{CFM}}(\theta) + \lambda\cdot \Lcal_{\mathrm{Cla}}(\theta)$ of a CFM-like loss $\Lcal_{\mathrm{CFM}}(\theta)$ and a loss $\Lcal_{\mathrm{Cla}}(\theta)$ dedicated to the change-of-class task.

The loss $\Lcal_{\mathrm{CFM}}(\theta)$ is obtained from \Cref{eq:cfm_loss_cond} with our coupling $\gamma$, and with the conditional path $p_{t,z}$ and the velocity $v^*_z=\xbf_1-\xbf_0$ of \Cref{eq:cond_path}:
\begin{align*}
&\Lcal_{\mathrm{CFM}}(\theta) =\!\!\!\!\!\!\! \EE_{\substack{t \sim \Ucal([0,1]),\\ ((y_0, y_1),(\xbf_0, \xbf_1)) \sim \gamma}} \!\!\!\!\!\left\| v_\theta(\xbf_t,t,y_0,y_1)-(\xbf_1{-}\xbf_0) \right\|^2\\
&=\!\!\!\!\! \EE_{\substack{t \sim \Ucal([0,1]),\\ ((y_0, y_1), B) \sim \gamma^n}}\sum_{(\xbf_0^{(i)}, \xbf_1^{(i)}) \in B}\!\!\!\!\!\! \tfrac{1}{n}\|v_\theta(\xbf_t^{(i)}\!\!,t,y_0,y_1){-}(\xbf_1^{(i)}{-}\xbf_0^{(i)})\|^2.
\end{align*}

\looseness=-1
Moreover, while \Cref{eq:counterfactual} requires generating realistic examples (according to $q_{y_1}$) with a minimal distance $d$, it also requires an example $\xbf_1$ such that $f(\xbf_1)=y_1$.
We add the loss $\Lcal_{\mathrm{Cla}}(\theta)$, weighted by $\lambda>0$, to further enforce a velocity field leading in the right direction. It is defined by
\begin{align*}
\Lcal_{\mathrm{Cla}}(\theta) &= \!\!\!\! \EE_{\substack{t \sim \Ucal([0,1]),\\ ((y_0, y_1),(\xbf_0, \xbf_1)) \sim \gamma}}\!\!\! \left[t\, \mathrm{CE} \!\left( f(\xbf_{t+\delta}), y_1 \right)\right]  = \!\! \EE_{\substack{t \sim \Ucal([0,1]),\\ ((y_0, y_1),B) \sim \gamma^n}}\sum_{(\xbf_0^{(i)}, \xbf_1^{(i)}) \in B} \tfrac{1}{n}\left[t\, \mathrm{CE} \!\left( f(\xbf_{t+\delta}^{(i)}), y_1 \right)\right],
\end{align*}
where $\mathrm{CE}$ is the classical multiclass cross-entropy loss, and $\xbf_{t+\delta}$ is defined as taking a small Euler step along the predicted velocity, \ie{}
$
    \xbf_{t+\delta} = \xbf_t + \delta\, v_\theta(\xbf_t,t,y_0,y_1).
$
The weighting factor $t$ progressively increases the influence of the classification objective: close to the source, the optimization is dominated by the flow matching objective, while near the target the classification constraint becomes stronger.

\begin{algorithm}[t]
\caption{$\mathrm{PairByEmbedding}$}
\label{alg:pair_by_embedding}
\textbf{Input}: VAE encoder $E$
\begin{algorithmic}[1]
\STATE Sample the labels $(y_0, y_1) \sim \Ucal(Y^2)$\label{line:train-labels}
\STATE Sample minibatches $B_0 \sim \Ucal(S_{y_0})^n$ and $B_1 \sim \Ucal(S_{y_1})^n$.\label{line:train-minibatch}
\FOR{$i=1,\dots,n$}\label{line:pair-loop}
    \STATE $\pi(i) \leftarrow \argmin_{j\in\{1,\dots,n\}} \left\| E(\xbf_0^{(i)}) - E(\xbf_1^{(j)}) \right\|_2$\label{line:pair-argmin}
\ENDFOR
\STATE $B \leftarrow \{(\xbf_0^{(i)}, \xbf_1^{(\pi(i))})\}_{i=1}^{n}$
\STATE \textbf{return} $((y_0,y_1),B)$\label{line:pair-return}
\end{algorithmic}
\end{algorithm}

\begin{algorithm}[t]
\caption{Training procedure of CounterFlow}
\label{alg:training}
\textbf{Input}: Source distribution $p_0$, target distribution $p_1$, flow network $v_\theta$, VAE encoder $E$, classifier $f$

\begin{algorithmic}[1]
\WHILE{Training}
    \STATE Sample $((y_0, y_1), B) \leftarrow \mathrm{PairByEmbedding}(E).$\label{line:train-pair}
    \FOR{$(\xbf_0^{(i)}, \xbf_1^{(i)}) \in B$}\label{line:train-loop}
        \STATE Sample $t\sim\mathcal{U}(0,1)$.\label{line:train-time}
        \STATE Compute \label{line:train-interp}
        $\xbf_t^{(i)} \leftarrow (1-t)\xbf_0^{(i)}+t\xbf_1^{(i)}$ \ and\ \  $\xbf_{t+\delta}^{(i)} \leftarrow \xbf_t^{(i)} + \delta\, v_\theta(\xbf_t^{(i)},t,y_0,y_1).$
        \STATE Compute the flow matching loss $\Lcal_{\mathrm{CFM}}^{(i)}(\theta) \leftarrow \left\| v_\theta(\xbf_t^{(i)},t,y_0,y_1)-(\xbf_1^{(i)}-\xbf_0^{(i)}) \right\|_2^2.$\label{line:train-fm}
        \STATE Compute the classification loss $\Lcal_{\mathrm{Cla}}^{(i)}(\theta) \leftarrow t\, \mathrm{CE} \!\left( f(\xbf_{t+\delta}^{(i)}), y_1 \right)$ \label{line:train-cla}
    \ENDFOR
\STATE Update $\theta$ using $\frac{1}{n}\sum_{i=1}^{n}\left[ \Lcal_{\mathrm{CFM}}^{(i)}(\theta) +\lambda\cdot\Lcal_{\mathrm{Cla}}^{(i)}(\theta)\right]$.\label{line:train-update}
\ENDWHILE
\STATE \textbf{return} $v_\theta$\label{line:train-return}
\end{algorithmic}
\end{algorithm}

\subsection{Training procedure (\Cref{alg:training})}

\looseness=-1
To minimize the loss function $\Lcal_{\mathrm{CFlow}}(\theta)$, we use the training loop described in \Cref{alg:training}.
A training step can be described in the following way.
We first sample $((y_0, y_1), B) \sim \gamma^n$ thanks to \Cref{alg:pair_by_embedding} (line~\ref{line:train-pair}), which corresponds to the labels $(y_0, y_1)$ and the minibatch $B$ with the pairs of examples established using nearest-neighbor matching in the latent space of a pretrained VAE encoder.
Then, for each matched pair $(\xbf_0^{(i)},\xbf_1^{(i)})\in B$, we sample $t\sim\mathcal{U}([0,1])$ (line~\ref{line:train-time}) and construct the interpolated samples (line~\ref{line:train-interp}).
Then, we compute the losses for $(\xbf_0^{(i)},\xbf_1^{(i)})\in B$ in lines~\ref{line:train-fm} and \ref{line:train-cla}.
After computing the loss on the minibatch, we update $v_{\theta}$ in line~\ref{line:train-update} using, \eg, Adam optimizer.

\subsection{Inference}

Once $v_\theta$ is trained, generating a counterfactual no longer requires any pairing, VAE encoding, or minibatch construction.
CounterFlow requires only an input sample $\xbf_0$ (with the prediction $y_0=f(\xbf_0)$) and a desired target class $y_1$.
The source class is obtained using the pretrained classifier, while the target class is either specified by the user or sampled uniformly among the remaining classes.
Starting from the input $\xbf_0$, we solve the learned ordinary differential equation defined in \Cref{eq:ode}, by integrating from $t=0$ to $t=1$, as shown in \Cref{eq:int-ode}.
The endpoint of the trajectory is returned as the generated counterfactual explanation $\xbf_1$.

\section{Experiments}
\label{sec:experiments}

We now turn to an empirical evaluation of CounterFlow against state-of-the-art counterfactual generation methods.
Note that our implementation is based on the \texttt{\href{https://github.com/atong01/conditional-flow-matching}{torchcfm}} library~\citep{tong2024improving}, which provides the Conditional Flow Matching framework used to train CounterFlow.
The source code to reproduce the experiments will be made publicly available at this \href{https://github.com/<anonymous>/CounterFlow}{link}.

\subsection{Setting}

\textbf{Datasets.}
We evaluate CounterFlow on two standard image classification benchmarks:
\textbf{\href{https://archive.ics.uci.edu/dataset/683}{MNIST}} and \textbf{\href{https://archive.ics.uci.edu/dataset/691}{CIFAR-10}}, MNIST consists of $70{,}000$ grayscale images of handwritten digits belonging to $10$ classes, with a spatial resolution of $28{\times}28$ pixels.
CIFAR-10 contains $60{,}000$ color images of size $32{\times}32$ pixels, evenly distributed among $10$ object categories.
These two datasets provide complementary evaluation settings, ranging from a simple handwritten digit classification task to a natural image classification benchmark.
For all experiments, we use the standard training and test splits provided with each dataset.
\Cref{sec:supp_datasets} discusses why we do not extend the comparison to larger benchmarks such as ImageNet.

\textbf{CounterFlow's inference.}
For all quantitative evaluations, CounterFlow is integrated via the Euler solver with 40 steps. This fixed-step solver provides a more deterministic evaluation protocol and allows a fair comparison of the computational cost across methods. For qualitative visualizations, we instead use the Dormand--Prince solver \citep{dormand1980family} (\texttt{dopri5}) to obtain smoother trajectories and higher-quality visual results.

\textbf{Methods.}
We compare CounterFlow\footnote{The hyperparameters of CounterFlow are listed in Appendix.} against three recent counterfactual generation methods: \href{https://github.com/khorrams/c3lt}{C3LT}~\citep{khorram2022cycle-consistent}, \href{https://github.com/ppegiosk/FastDiME_CelebA}{FastDiME}~\citep{weng2024fast}, and \href{https://github.com/caicairay/LeapFactual}{LeapFactual}~\citep{cao2025leapfactual}, where we use the official implementations.
Among diffusion-guidance methods, we retain only FastDiME as a representative baseline: DiME and ACE are prohibitively expensive to run at scale due to their inner-loop or per-step attack computations; DVCE requires an adversarially robust classifier, a constraint that our method does not rely on and that we therefore do not impose in our experiments; and MaskDiME's masking mechanism restricts edits to a localized region, which conflicts with our evaluation setting where the change may need to affect the whole image.
Beyond this choice, neither of the two remaining baselines, namely LeapFactual and FastDiME, is applicable to both datasets.
LeapFactual relies on a compact VAE latent space to remain tractable, which does not scale to CIFAR-10 without dominant reconstruction artifacts and a much costlier latent flow, so we evaluate it only on MNIST.
Conversely, FastDiME depends on pretrained OpenAI diffusion models built for natural images, so we evaluate it only on CIFAR-10.

\textbf{Models to be explained.}
\looseness=-1
To ensure a fair comparison, all methods rely on the same pretrained image classifier for a given dataset.
For MNIST, we trained a simple convolutional neural network achieving an accuracy of $99.9\%$ on the test set.
For CIFAR-10, we use a pretrained RepVGG-A2 classifier \citep{ding2021repvgg} with publicly available weights from the \texttt{pytorch-cifar-models} repository (distributed through the official PyTorch Hub interface), achieving a test accuracy of $94.8\%$.
Both classifiers are frozen during training of counterfactual models and are used consistently for counterfactual generation and metrics computation across all competing methods.

\textbf{Metrics.}
We evaluate the counterfactual explanations along four axes: \emph{effectiveness} (reaching $f(\xbf_1)=y_1$), \emph{proximity to the source} (a distribution-level counterpart of the distance $d(\xbf_1,\xbf_0)$ minimized per example in \Cref{eq:counterfactual}, measured in a semantic feature space rather than in a pixel space), \emph{realism} (the plausibility constraint $q_{y_1}(\xbf_1)\ge\varepsilon$, also from \Cref{eq:counterfactual}), and \emph{computational efficiency}.
We operationalize the evaluation of these axes through the following five metrics.
\begin{itemize}
    \item \textbf{Conditional Validity} measures the proportion (in \%) of generated counterfactuals $x_1$ that are classified as the desired target class $y_1$, \ie, whether we have $f(\xbf_1)=y_1$.
    \item \textbf{FID$_{\mathrm{t}}$} computes the Fréchet Inception Distance between the generated counterfactuals and real samples from the target class, quantifying \emph{realism}: lower values indicate that the generated samples better match the distribution of the desired class.
    \item \textbf{FID$_{\mathrm{s}}$} computes the Fréchet Inception Distance between the generated counterfactuals and the original source class, measuring \emph{proximity to the source}: lower values indicate that the counterfactuals remain closer to the original data distribution, suggesting that fewer semantic changes were applied to reach the target class.
    \item \textbf{FLOPs / sample} measures the computational cost required to generate a single counterfactual explanation.
    It includes the complete inference pipeline, \ie, all neural network evaluations required by the ODE solver.
    This metric provides an implementation-independent estimate of inference complexity.
    \item \textbf{Model Size} reports the total number of trainable parameters of each method, allowing comparison of their memory footprint and model complexity.
\end{itemize}

Following the standard definition of the Fréchet Inception Distance \citep[FID,][]{heusel2017fid}, feature statistics for FID$_{\mathrm{t}}$ and FID$_{\mathrm{s}}$ are extracted from pretrained image classification networks.
For CIFAR-10, we use the pretrained Inception-V3 network available in \texttt{torchvision}.
Since ImageNet-pretrained models are not well suited for handwritten digit recognition, for MNIST we instead compute features using the pretrained convolutional classifier employed throughout our experiments.
FID is then computed between the Gaussian statistics of the extracted features from the generated counterfactuals and those of the corresponding source or target class, \ie, FID$_{\mathrm{t}}$ on the target class and FID$_{\mathrm{s}}$ on the source class.
Note that, since these statistics are estimated from the subset of test examples belonging to a single class, rather than from the full test set as in standard, unconditional FID computations, the reported values are higher than those typically observed in unconditional image generation.
We therefore focus on relative comparisons between methods under the same evaluation protocol rather than on the absolute reported numbers.

\begin{table}[t!]
    \centering
    \caption{(a) Quantitative results and (b) Ablation of the pairing used to train CounterFlow: latent-space nearest neighbor (\Cref{alg:pair_by_embedding}) vs.\ random. Higher conditional validity (CV) is better, while lower FID scores indicate that the generated counterfactuals are closer to the target or source distributions. Lower FLOPs/sample and model size (MS) are preferred for computational efficiency.}
    \setlength{\tabcolsep}{4pt}
    \begin{subtable}[b]{0.60\textwidth}
    \centering
    \begin{tabular}{p{1.7cm}p{0.6cm}p{0.9cm}p{1.1cm}p{1.4cm}p{1.4cm}}
        \toprule
        Method & \textbf{CV}$\uparrow$ & \textbf{FID$_{\mathrm{t}}$}$\downarrow$ & \textbf{FID$_{\mathrm{s}}$}$\downarrow$ & \textbf{FLOPs}$\downarrow$ & \textbf{MS}$\downarrow$ \\
        \midrule
        \multicolumn{6}{c}{\textbf{MNIST}} \\
        \midrule
        C3LT & 99.0 & 187.89 & 4310.01 & 223.86M & 480.81M \\
        LeapFactual & 98.6 & 217.08 & \textbf{2625.52} & \textbf{0.86M} & \textbf{56.95K} \\
        CounterFlow & \textbf{99.6} & \textbf{105.70} & 3224.54 & 174.66M & 4.64M \\
        \midrule
        \multicolumn{6}{c}{\textbf{CIFAR10}} \\
        \midrule
        C3LT & 99.0 & 115.04 & 153.41 & 3.93G & 481.33M \\
        FastDiME & 16.0 & 159.86 & 175.64 & 1586.63G & 52.55M \\
        CounterFlow & \textbf{100.0} & \textbf{62.54} & \textbf{117.03} & \textbf{1.73G} & \textbf{39.36M} \\
        \bottomrule
    \end{tabular}
    \caption{Quantitative results}
    \label{tab:results}
    \end{subtable}\begin{subtable}[b]{0.40\textwidth}
    \centering
    \begin{tabular}{p{1.0cm}p{0.6cm}p{0.9cm}p{1.0cm}}
        \toprule
        Pairing & \textbf{CV}$\uparrow$ & \textbf{FID$_{\mathrm{t}}$}$\downarrow$ & \textbf{FID$_{\mathrm{s}}$}$\downarrow$ \\
        \midrule
        \multicolumn{4}{c}{\textbf{MNIST}} \\
        \midrule
        Random & 99.4 & 111.98 & 3634.20 \\
        Latent & \textbf{99.6} & \textbf{105.70} & \textbf{3224.54} \\
        \midrule
        \multicolumn{4}{c}{\textbf{CIFAR10}} \\
        \midrule
        Random & 99.9 & 63.27 & 125.07 \\
        Latent & \textbf{100.0} & \textbf{62.54} & \textbf{117.03} \\
        \bottomrule
    \end{tabular}
    \caption{Pairing used to train CounterFlow}
    \label{tab:ablation_pairing}
    \end{subtable}
\end{table}

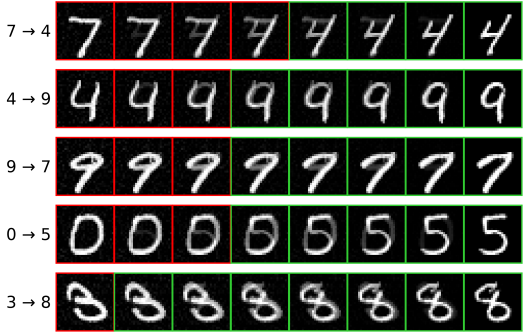
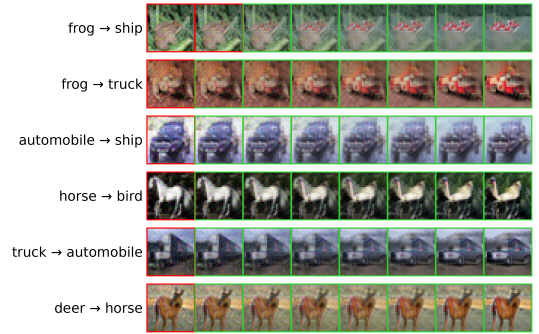
\begin{figure*}[t]
    \centering
    \begin{subfigure}[t]{0.44\linewidth}
        \centering
        \includegraphics[width=\linewidth]{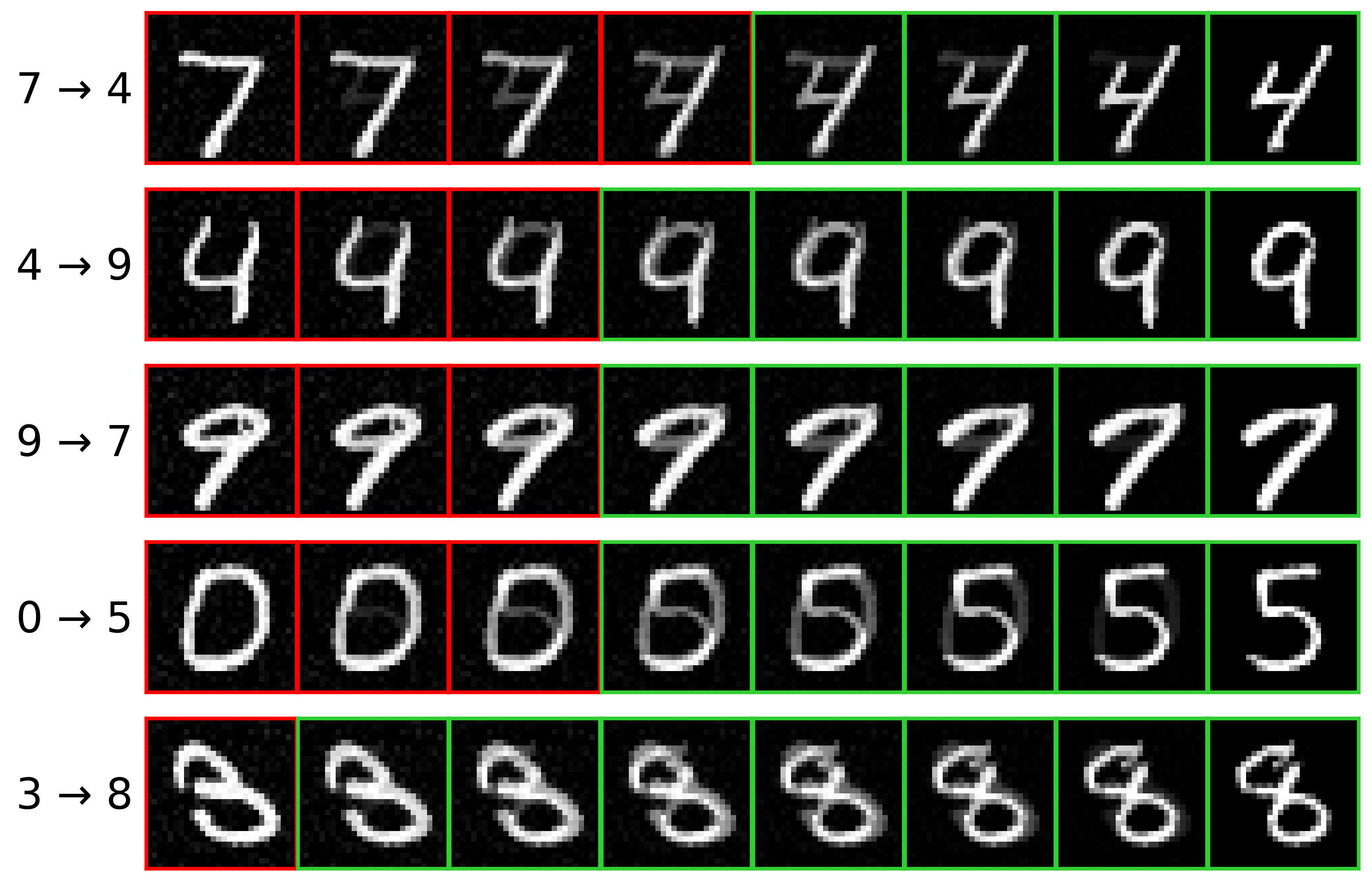}
        \subcaption{MNIST}
        \label{fig:traj:mnist}
    \end{subfigure}
    \hfill
    \begin{minipage}[t]{0pt}
        \begin{tikzpicture}
            \draw[dashed, gray] (0,0) -- (0,-3.9);
        \end{tikzpicture}
    \end{minipage}
    \hfill
    \begin{subfigure}[t]{0.44\linewidth}
        \centering
        \includegraphics[width=\linewidth]{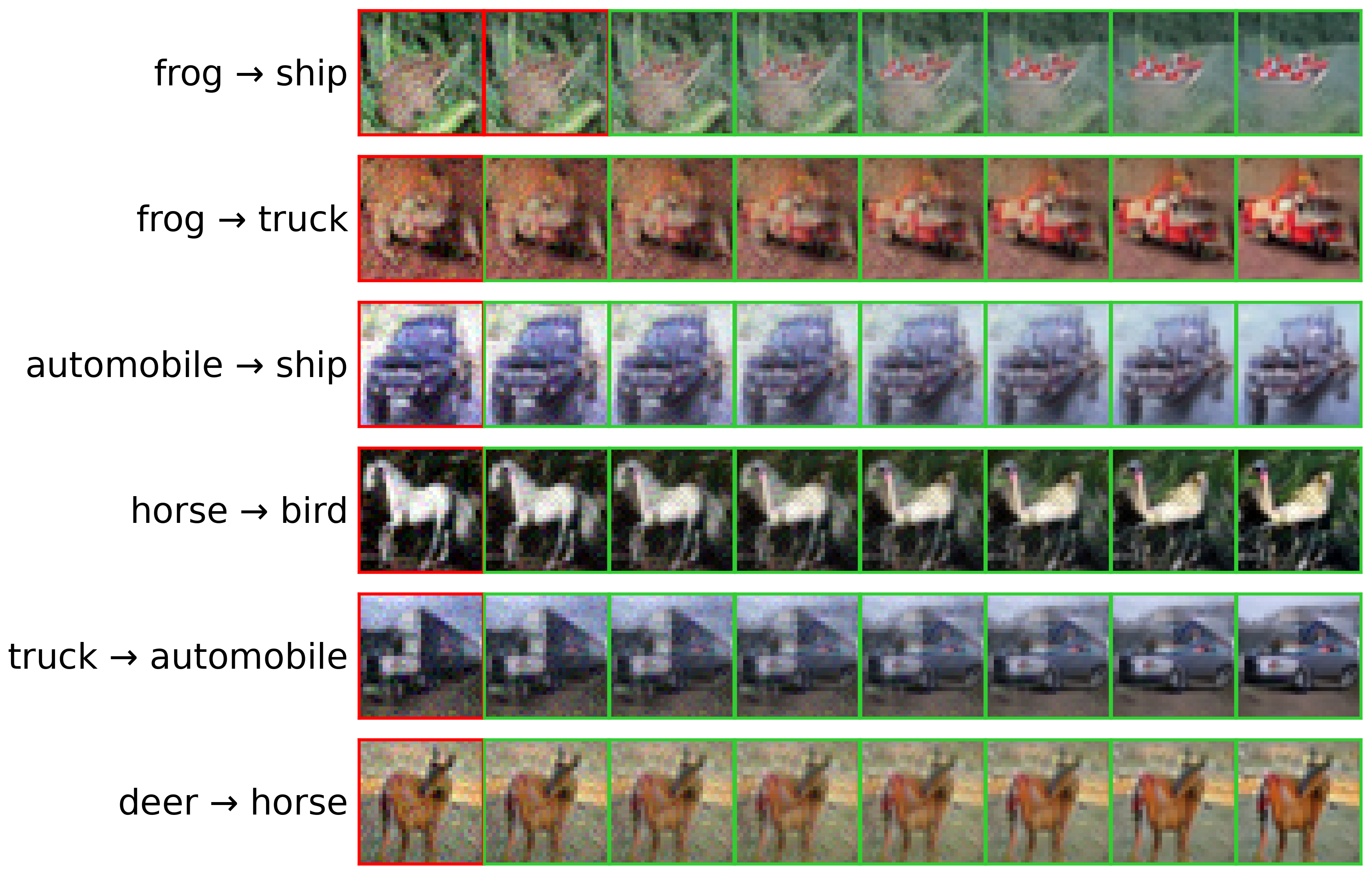}
        \subcaption{CIFAR10}
        \label{fig:traj:cifar}
    \end{subfigure}
    \caption{%
        CounterFlow trajectories for five (MNIST) and six (CIFAR10) samples, from source to target class (labeled for CIFAR10). Border color indicates the classifier's prediction along the path: \textcolor{red}{red} for the source class, \textcolor{green!50!black}{green} for the target class.
    }
    \label{fig:traj_qualitative}
\end{figure*}

\subsection{Quantitative Results}

\Cref{tab:results} reports the quantitative comparison between CounterFlow and representative counterfactual generation methods on MNIST and CIFAR-10; a radar-plot visualization is provided in \Cref{fig:radar_results} in the appendix.

\textbf{MNIST.} CounterFlow reaches the highest conditional validity ($99.6\%$) and the lowest target FID while remaining close to the source, confirming that the learned conditional transport reliably reaches the target class with a single ODE integration.
LeapFactual stays closer to the source but has the highest target FID, reflecting a more conservative transformation.
C3LT performs worst overall and requires by far the largest model, due to its per-class-pair VAE, GAN, and mapping networks.

\textbf{CIFAR-10.} CounterFlow is best on all five metrics: it reaches $100\%$ conditional validity, nearly halves C3LT's target FID while also lowering its source FID by a quarter, and needs roughly half the inference FLOPs and an order of magnitude fewer parameters. Improving both FIDs at once is the signature of a near-minimal transport: counterfactuals that fall within the target class without drifting away from the source image. FastDiME reaches only $16\%$ conditional validity despite a diffusion backbone $900\times$ more expensive per sample, and has the worst FID on both axes; with so few samples switching class, these scores reflect generic realism rather than class transfer.

\looseness=-1
Overall, CounterFlow achieves state-of-the-art counterfactual validity and the best realism/proximity trade-off on CIFAR-10, while substantially reducing inference complexity compared with prior flow- and diffusion-based approaches.

\textbf{Ablation.} \Cref{tab:ablation_pairing} shows that replacing latent-space pairing with random pairing consistently lowers conditional validity and raises both target and source FID on both datasets, mostly through the source FID on CIFAR-10. Pairing semantically similar examples yields simpler transport trajectories that are easier for the conditional velocity field to approximate, whereas random pairing often associates visually unrelated samples, leading to harder-to-learn trajectories and lower-quality counterfactuals. \Cref{sec:supp_ablations} further ablates $\lambda$, the conditioning context and the explained classifier.

\subsection{Qualitative Results}

\textbf{Style Preservation.} \Cref{fig:mnist_cifar_qualitative}
shows five source/counterfactual pairs per dataset. On MNIST, CounterFlow preserves writer-specific style, such as stroke thickness and slant -- \eg, the $4{\to}5$ transition retains the source's thin strokes -- and keeps the digit's orientation, rather than collapsing to a generic class prototype. On CIFAR10, the same near-minimal-path behavior appears at the scene level: the background stays untouched and only the foreground object changes, often preserving its color palette and orientation while altering just the class-relevant structure.

\looseness=-1
\textbf{Trajectories.} \Cref{fig:traj_qualitative} visualizes the full integration trajectory for five counterfactuals per dataset, with the border color marking the classifier's prediction at each step (\textcolor{red}{red} for the source class, \textcolor{green!50!black}{green} for the target class). In most examples the color flips early, showing that CounterFlow reaches the target class within a small fraction of the path and spends the remaining steps refining the sample within that class rather than driving the decision itself, \ie, most of the trajectory makes the class change plausible rather than induces it. Intermediate frames stay visually close to both endpoints rather than passing through a degraded intermediate state, confirming at a finer granularity that CounterFlow produces minimal, semantically meaningful edits rather than an unconstrained image-to-image mapping.
See \Cref{fig:additional_trajectories_mnist,fig:additional_trajectories_cifar10} in \Cref{sec:supp_additional_results} for additional CounterFlow 
trajectories.

\section{Discussion}
\label{sec:discussion}

Existing image counterfactual generation methods often require heavy generative backbones, including diffusion models or multiple class-specific generative networks, making their deployment computationally demanding. CounterFlow demonstrates that Conditional Flow Matching provides a significantly simpler alternative. By learning a single target-conditioned transport field and requiring only one ODE integration at inference, our method achieves competitive or superior counterfactual quality while substantially reducing computational complexity. We believe this constitutes an important step toward making counterfactual explanations practical for interactive decision-support systems, where explanations must be generated quickly and consistently.

Despite these encouraging results, several limitations remain. First, our experiments are restricted to relatively low-resolution image datasets. Evaluating CounterFlow on larger-scale benchmarks such as ImageNet or high-resolution generative datasets would provide further evidence of its scalability. Second, our latent-space pairing strategy currently relies on nearest-neighbor retrieval using a pretrained encoder, which introduces an additional preprocessing step. Finally, while Conditional Flow Matching substantially reduces inference complexity compared to diffusion models, solving an ODE remains more expensive than a single feed-forward network.

\section{Conclusion}
\label{sec:conclusion}

We presented CounterFlow, a Conditional Flow Matching framework for visual counterfactual explanation generation. Unlike previous approaches relying on GANs or diffusion models, CounterFlow directly learns a target-conditioned transport field between image classes in pixel space. Experiments on MNIST and CIFAR-10 demonstrate that the proposed method achieves state-of-the-art counterfactual validity while remaining competitive in image realism and substantially reducing computational requirements.

Beyond counterfactual explanations, CounterFlow illustrates that target-conditioned flow matching can efficiently learn semantic transformations directly between complex data distributions, without relying on an intermediate noise distribution. Future work will investigate its application to larger-scale datasets and other data transformation tasks, including domain adaptation, semantic image editing, and robust model analysis.

\subsection*{AI use statement}

We used generative AI tools (LLM-based coding assistants) to help implement and debug parts of the experimental code, and to polish the phrasing of some sentences in this manuscript. We did not use generative AI tools to generate research ideas, to design the method, or to produce or interpret experimental results. All AI-assisted code and text were reviewed, tested, and validated by the authors, who take full responsibility for the final content of this work, including text, claims, and artifacts produced with the aid of generative AI.

\subsection*{Ethics statement}

This work introduces a method for generating visual counterfactual explanations of image classifiers, evaluated on the standard MNIST and CIFAR-10 benchmarks, which do not contain personal or sensitive data. As with other counterfactual-explanation and image-editing methods, the underlying generative technology could in principle be misused to produce misleading edited images of real people; we mitigate this risk by restricting our experiments to non-human-identifying benchmark datasets and by releasing the method as a tool for model interpretability rather than for unconstrained image generation.

\subsection*{Reproducibility statement}

We describe the CounterFlow training and inference procedures in full in \Cref{sec:counter_flow}, including the pairing algorithm (\Cref{alg:pair_by_embedding}) and the training loop (\Cref{alg:training}). All training hyperparameters for both datasets are listed in \Cref{tab:hyperparams} in the appendix. We use only the standard, publicly available MNIST and CIFAR-10 datasets and publicly available pretrained classifiers (\Cref{sec:experiments}). Our implementation builds on the open-source \texttt{torchcfm} library, and we will release the source code needed to reproduce all experiments upon publication.

\bibliography{main}

@article{guo2026maskdime,
  author       = {Changlu Guo and
                  Anders Nymark Christensen and
                  Anders Bjorholm Dahl and
                  Morten Rieger Hannemose},
  title        = {{MaskDiME: Adaptive Masked Diffusion for Precise and Efficient Visual
                  Counterfactual Explanations}},
  journal      = {arXiv},
  volume       = {abs/2602.18792},
  year         = {2026},
}

@inproceedings{cao2025leapfactual,
  author       = {Zhuo Cao and
                  Xuan Zhao and
                  Lena Krieger and
                  Hanno Scharr and
                  Ira Assent},
  title        = {{LeapFactual: Reliable Visual Counterfactual Explanation Using Conditional
                  Flow Matching}},
  booktitle    = {Advances in Neural Information Processing Systems},
  year         = {2025}
}

@article{tong2024improving,
  author       = {Alexander Tong and
                  Kilian Fatras and
                  Nikolay Malkin and
                  Guillaume Huguet and
                  Yanlei Zhang and
                  Jarrid Rector{-}Brooks and
                  Guy Wolf and
                  Yoshua Bengio},
  title        = {{Improving and generalizing flow-based generative models with minibatch
                  optimal transport}},
  journal      = {Transactions on Machine Learning Research},
  year         = {2024}
}

@inproceedings{weng2024fast,
  author       = {Nina Weng and
                  Paraskevas Pegios and
                  Eike Petersen and
                  Aasa Feragen and
                  Siavash Arjomand Bigdeli},
  title        = {{Fast Diffusion-Based Counterfactuals for Shortcut Removal and Generation}},
  booktitle    = {European Conference on Computer Vision},
  year         = {2024}
}

@article{jeanneret2024diffusion,
  author       = {Guillaume Jeanneret and
                  Lo{\"{\i}}c Simon and
                  Fr{\'{e}}d{\'{e}}ric Jurie},
  title        = {{Diffusion Models for Counterfactual Explanations}},
  journal      = {Computer Vision and Image Understanding},
  year         = {2024}
}

@inproceedings{liu2023audioldm,
  author       = {Haohe Liu and
                  Zehua Chen and
                  Yi Yuan and
                  Xinhao Mei and
                  Xubo Liu and
                  Danilo Mandic and
                  Wenwu Wang and
                  Mark D. Plumbley},
  title        = {{AudioLDM: Text-to-Audio Generation with Latent Diffusion Models}},
  booktitle    = {International Conference on Machine Learning},
  year         = {2023}
}

@inproceedings{albergo2023building,
  author       = {Michael S. Albergo and
                  Eric Vanden{-}Eijnden},
  title        = {{Building Normalizing Flows with Stochastic Interpolants}},
  booktitle    = {International Conference on Learning Representations},
  year         = {2023}
}

@inproceedings{liu2023flow,
  author       = {Xingchao Liu and
                  Chengyue Gong and
                  Qiang Liu},
  title        = {{Flow Straight and Fast: Learning to Generate and Transfer Data with
                  Rectified Flow}},
  booktitle    = {International Conference on Learning Representations},
  year         = {2023}
}

@inproceedings{lipman2023flow,
  author       = {Yaron Lipman and
                  Ricky T. Q. Chen and
                  Heli Ben{-}Hamu and
                  Maximilian Nickel and
                  Matthew Le},
  title        = {{Flow Matching for Generative Modeling}},
  booktitle    = {International Conference on Learning Representations},
  year         = {2023}
}

@inproceedings{jeanneret2023adversarial,
  author       = {Guillaume Jeanneret and
                  Lo{\"{\i}}c Simon and
                  Fr{\'{e}}d{\'{e}}ric Jurie},
  title        = {{Adversarial Counterfactual Visual Explanations}},
  booktitle    = {IEEE/CVF Conference on Computer Vision and Pattern Recognition},
  year         = {2023}
}

@article{dwivedi2023explainable,
  author       = {Rudresh Dwivedi and
                  Devam Dave and
                  Het Naik and
                  Smiti Singhal and
                  Rana Omer and
                  Pankesh Patel and
                  Bin Qian and
                  Zhenyu Wen and
                  Tejal Shah and
                  Graham Morgan and
                  Rajiv Ranjan},
  title        = {{Explainable AI (XAI): Core Ideas, Techniques, and Solutions}},
  journal      = {ACM Computing Surveys},
  year         = {2023}
}

@inproceedings{augustin2022diffusion,
  author       = {Maximilian Augustin and
                  Valentyn Boreiko and
                  Francesco Croce and
                  Matthias Hein},
  title        = {{Diffusion Visual Counterfactual Explanations}},
  booktitle    = {Advances in Neural Information Processing Systems},
  year         = {2022}
}

@inproceedings{khorram2022cycle-consistent,
  author       = {Saeed Khorram and
                  Li Fuxin},
  title        = {{Cycle-Consistent Counterfactuals by Latent Transformations}},
  booktitle    = {IEEE/CVF Conference on Computer Vision and Pattern Recognition},
  year         = {2022}
}

@inproceedings{rombach2022high-resolution,
  author       = {Robin Rombach and
                  Andreas Blattmann and
                  Dominik Lorenz and
                  Patrick Esser and
                  Bj{\"o}rn Ommer},
  title        = {{High-Resolution Image Synthesis With Latent Diffusion Models}},
  booktitle    = {IEEE/CVF Conference on Computer Vision and Pattern Recognition},
  year         = {2022}
}

@inproceedings{ho2022video,
  author       = {Jonathan Ho and
                  Tim Salimans and
                  Alexey Gritsenko and
                  William Chan and
                  Mohammad Norouzi and
                  David J. Fleet},
  title        = {{Video diffusion models}},
  booktitle    = {Advances in Neural Information Processing Systems},
  year         = {2022}
}

@inproceedings{gowal2021improving,
  author       = {Sven Gowal and
                  Sylvestre{-}Alvise Rebuffi and
                  Olivia Wiles and
                  Florian Stimberg and
                  Dan Andrei Calian and
                  Timothy A. Mann},
  title        = {{Improving Robustness using Generated Data}},
  booktitle    = {Advances in Neural Information Processing Systems},
  year         = {2021}
}

@inproceedings{ding2021repvgg,
  author       = {Xiaohan Ding and
                  Xiangyu Zhang and
                  Ningning Ma and
                  Jungong Han and
                  Guiguang Ding and
                  Jian Sun},
  title        = {{Repvgg: Making vgg-style convnets great again}},
  booktitle    = {IEEE/CVF Conference on Computer Vision and Pattern Recognition},
  year         = {2021}
}

@inproceedings{keane2020good,
  author       = {Mark T Keane and
                  Barry Smyth},
  title        = {{Good Counterfactuals and Where to Find Them: A Case-Based Technique for Generating Counterfactuals for Explainable AI (XAI)}},
  booktitle    = {International Conference on Case-Based Reasoning},
  year         = {2020}
}

@inproceedings{goyal2019counterfactual,
  author       = {Yash Goyal and
                  Ziyan Wu and
                  Jan Ernst and
                  Dhruv Batra and
                  Devi Parikh and
                  Stefan Lee},
  title        = {{Counterfactual Visual Explanations}},
  booktitle    = {International Conference on Machine Learning},
  year         = {2019}
}

@article{artelt2019on,
  author       = {Andr{\'{e}} Artelt and
                  Barbara Hammer},
  title        = {{On the computation of counterfactual explanations - {A} survey}},
  journal      = {arXiv},
  volume       = {abs/1911.07749},
  year         = {2019}
}

@inproceedings{chen2018neural,
  author       = {Tian Qi Chen and
                  Yulia Rubanova and
                  Jesse Bettencourt and
                  David Duvenaud},
  title        = {{Neural Ordinary Differential Equations}},
  booktitle    = {Advances in Neural Information Processing Systems},
  year         = {2018}
}

@article{wachter2017counterfactual,
  author       = {Sandra Wachter and
                  Brent D. Mittelstadt and
                  Chris Russell},
  title        = {{Counterfactual Explanations without Opening the Black Box: Automated
                  Decisions and the {GDPR}}},
  journal      = {arXiv},
  volume       = {abs/1711.00399},
  year         = {2017}
}

@inproceedings{artelt2020convex,
  author       = {Andr{\'{e}} Artelt and
                  Barbara Hammer},
  title        = {{Convex Density Constraints for Computing Plausible Counterfactual Explanations}},
  booktitle    = {International Conference on Artificial Neural Networks},
  year         = {2020}
}

@inproceedings{heusel2017fid,
  title={{GANs} trained by a two time-scale update rule converge to a local {Nash} equilibrium},
  author={Heusel, Martin and Ramsauer, Hubert and Unterthiner, Thomas and Nessler, Bernhard and Hochreiter, Sepp},
  booktitle={Advances in Neural Information Processing Systems},
  year={2017}
}

@article{dormand1980family,
  author    = {J. R. Dormand and
               P. J. Prince},
  title     = {{A Family of Embedded Runge-Kutta Formulae}},
  journal   = {Journal of Computational and Applied Mathematics},
  year      = {1980}
}
\bibliographystyle{plainnat}

\appendix
\section{Appendix}

\subsection{Implementation Details}
\label{sec:supp_hyperparams}

We provide the full training configuration for both datasets in \Cref{tab:hyperparams} to facilitate reproducibility. Both models are trained with the Adam optimizer and an exponential moving average (EMA) of the weights, using a linear warmup schedule for both the learning rate and the classifier-guidance loss weight $\lambda$.

\begin{table}[h]
    \centering
    \caption{Training hyperparameters for MNIST and CIFAR10.}
    \label{tab:hyperparams}
    \setlength{\tabcolsep}{4pt}
    \begin{tabular}{lcc}
        \toprule
        Hyperparameter & MNIST & CIFAR10 \\
        \midrule
        Training steps & 200{,}000 & 400{,}000 \\
        Warmup steps & 5{,}000 & 5{,}000 \\
        Minibatch size & 256 & 256 \\
        Optimizer & Adam & Adam \\
        Learning rate & $1\times10^{-4}$ & $1\times10^{-4}$ \\
        EMA decay & 0.9999 & 0.9999 \\
        Validation ratio & 0.2 & 0.2 \\
        Seed & 1 & 1 \\
        \midrule
        $\lambda$ & 0.1 & 0.1 \\
        $\lambda$ warmup steps & 10{,}000 & 10{,}000 \\
        Noise level & 0.1 & 0.1 \\
        $\delta$ & 0.05 & 0.05 \\
        \midrule
        Number of residual blocks in UNet & 1 & 2 \\
        Conditioning MLP & \ding{55} & \ding{51} \\
        \midrule
        Image size ($C \times H \times W$) & $1\times28\times28$ & $3\times32\times32$ \\
        Number of classes & 10 & 10 \\
        \bottomrule
    \end{tabular}
\end{table}

\subsection{Radar plots}

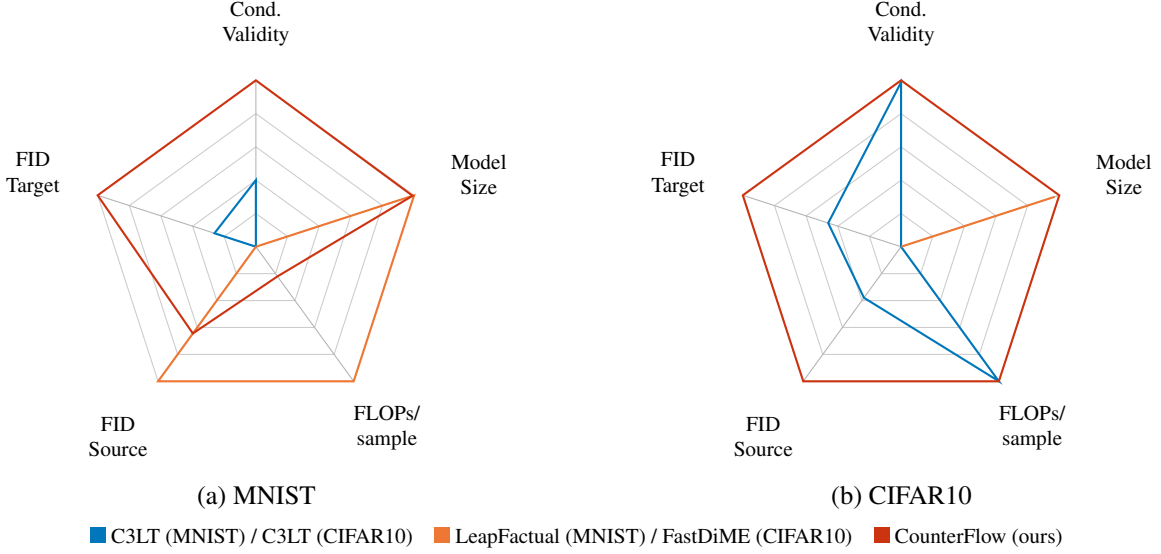
\begin{figure*}[h]
    \centering
    \begin{subfigure}[h]{0.46\linewidth}
        \centering
        \begin{tikzpicture}[scale=1]
            \def\R{2.2}
            \radaraxis{\R}

            % C3LT (blue)
            \draw[thick, blue, mark=*, mark size=1.5pt]
                (90:0.4*\R) -- (162:0.262*\R) -- (234:0*\R)
                -- (306:0*\R) -- (18:0*\R) -- cycle;

            % LeapFactual (orange)
            \draw[thick, orange, mark=*, mark size=1.5pt]
                (90:0*\R) -- (162:0*\R) -- (234:1*\R)
                -- (306:1*\R) -- (18:1*\R) -- cycle;

            % CounterFlow (ours, red)
            \draw[thick, red, mark=*, mark size=1.5pt]
                (90:1*\R) -- (162:1*\R) -- (234:0.6445*\R)
                -- (306:0.2206*\R) -- (18:0.9905*\R) -- cycle;
        \end{tikzpicture}
        \subcaption{MNIST}
        \label{fig:radar:mnist}
    \end{subfigure}
    \hfill
    \begin{subfigure}[h]{0.46\linewidth}
        \centering
        \begin{tikzpicture}[scale=1]
            \def\R{2.2}
            \radaraxis{\R}

            % C3LT (blue)
            \draw[thick, blue, mark=*, mark size=1.5pt]
                (90:0.988*\R) -- (162:0.4605*\R) -- (234:0.3793*\R)
                -- (306:0.9986*\R) -- (18:0*\R) -- cycle;

            % FastDiME (orange)
            \draw[thick, orange, mark=*, mark size=1.5pt]
                (90:0*\R) -- (162:0*\R) -- (234:0*\R)
                -- (306:0*\R) -- (18:0.9702*\R) -- cycle;

            % CounterFlow (ours, red)
            \draw[thick, red, mark=*, mark size=1.5pt]
                (90:1*\R) -- (162:1*\R) -- (234:1*\R)
                -- (306:1*\R) -- (18:1*\R) -- cycle;
        \end{tikzpicture}
        \subcaption{CIFAR10}
        \label{fig:radar:cifar}
    \end{subfigure}

    \vspace{4pt}
    \centering
    {\scriptsize
    \textcolor{blue}{\rule{6pt}{6pt}}~C3LT (MNIST) / C3LT (CIFAR10)\quad
    \textcolor{orange}{\rule{6pt}{6pt}}~LeapFactual (MNIST) / FastDiME (CIFAR10)\quad
    \textcolor{red}{\rule{6pt}{6pt}}~CounterFlow (ours)
    }

    \caption{%
        Radar comparison of all five metrics from \Cref{tab:results}, per dataset. Each axis is min-max normalized within its dataset so that the outer edge always corresponds to the best value across methods, regardless of whether the underlying metric is originally better when higher or lower. CounterFlow reaches the outer edge on every axis on CIFAR-10 and is competitive or best on most axes on MNIST, whereas baselines tend to excel on a single axis at the expense of others.
    }
    \label{fig:radar_results}
\end{figure*}

\subsection{Ablation studies on CIFAR-10}
\label{sec:supp_ablations}

\Cref{tab:ablation_lambda,tab:ablation_context,tab:ablation_classifier} isolate the three design choices of CounterFlow on CIFAR-10.
Every run follows the configuration of \Cref{tab:hyperparams} and modifies a single factor: models are trained for $400{,}000$ steps and evaluated on the full CIFAR-10 test set with the Euler solver and $40$ steps.
The rows marked \emph{(default)} all correspond to the configuration reported in \Cref{tab:results}.
Rows marked $\dagger$ have a conditional validity low enough that their FID scores are computed over a mixture of successful and failed counterfactuals.
Since a counterfactual that stays in the source class trivially obtains a low FID$_{\mathrm{s}}$ and a high FID$_{\mathrm{t}}$, these values must not be read as a quality signal.

\textbf{Weight $\lambda$ of the classification loss.}
\Cref{tab:ablation_lambda} varies the weight of $\Lcal_{\mathrm{Cla}}$ in $\Lcal_{\mathrm{CFlow}}(\theta) = \Lcal_{\mathrm{CFM}}(\theta) + \lambda\cdot\Lcal_{\mathrm{Cla}}(\theta)$.
The two terms are complementary, and neither is sufficient on its own.
Removing the classification term ($\lambda=0$) leaves a purely generative transport that reaches only $68.9\%$ conditional validity: matching the flow between the two class distributions does not guarantee that each individual sample crosses the decision boundary of $f$, and its lower FID$_{\mathrm{s}}$ merely reflects that about a third of the samples never leave the source class.
Conversely, keeping only the classification term ($\lambda=1$, no $\Lcal_{\mathrm{CFM}}$) collapses: with nothing anchoring the velocity field to the source--target displacement, the model drifts off the data manifold, as the FID scores above $400$ make explicit, and it does not even reach a high validity ($48.9\%$).
Between these two extremes, every $\lambda \in [0.1, 0.5]$ saturates conditional validity at $100\%$, so the remaining effect of $\lambda$ is a monotonic degradation of both FID scores, from $62.38$ to $74.01$ on the target and from $118.23$ to $121.82$ on the source.
We therefore use $\lambda=0.1$, the smallest value that already saturates validity.

\begin{table}[H]
    \centering
    \caption{Effect of the weight $\lambda$ of the classification loss $\Lcal_{\mathrm{Cla}}$ on CIFAR-10.}
    \label{tab:ablation_lambda}
    \setlength{\tabcolsep}{4pt}
    \begin{tabular}{lccc}
        \toprule
        Ablated factor & \textbf{CV}$\uparrow$ & \textbf{FID$_{\mathrm{t}}$}$\downarrow$ & \textbf{FID$_{\mathrm{s}}$}$\downarrow$ \\
        \midrule
        $\lambda=0$ (no $\Lcal_{\mathrm{Cla}}$)$^{\dagger}$ & 68.9 & 67.64 & 112.94 \\
        $\lambda=0.1$ \emph{(default)} & 100.0 & 62.54 & 117.03 \\
        $\lambda=0.2$ & 100.0 & 66.78 & 119.77 \\
        $\lambda=0.5$ & 100.0 & 74.01 & 121.82 \\
        $\lambda=1$ (no $\Lcal_{\mathrm{CFM}}$)$^{\dagger}$ & 48.9 & 444.88 & 406.14 \\
        \bottomrule
    \end{tabular}
\end{table}

\textbf{Conditioning context.}
\Cref{tab:ablation_context} removes the context $c=(y_0,y_1)$ from the velocity field, \ie, it trains $v_\theta(\xbf_t,t)$ on the same latent-paired minibatches, but without indicating which transition is being realized.
Conditional validity collapses from $100\%$ to $8.6\%$: deprived of the context, the model can only fit the average displacement over all source--target pairs, which carries no information about the requested transition.
The FID scores confirm that the resulting field is close to the identity, since the source FID falls to $83.24$, well below the default $118.23$, precisely because the generated images barely leave their source.
Conditioning on the pair $(y_0,y_1)$ is thus not an implementation detail, but the component that turns a generic flow into a controllable, target-specific transport.

\begin{table}[H]
    \centering
    \caption{Effect of the context $c$ conditioning the velocity field $v_\theta$ on CIFAR-10.}
    \label{tab:ablation_context}
    \setlength{\tabcolsep}{4pt}
    \begin{tabular}{lccc}
        \toprule
        Ablated factor & \textbf{CV}$\uparrow$ & \textbf{FID$_{\mathrm{t}}$}$\downarrow$ & \textbf{FID$_{\mathrm{s}}$}$\downarrow$ \\
        \midrule
        $c=(y_0,y_1)$ \emph{(default)} & 100.0 & 62.54 & 117.03 \\
        No context$^{\dagger}$ & 8.6 & 121.72 & 83.24 \\
        \bottomrule
    \end{tabular}
\end{table}

\textbf{Explained classifier.}
A counterfactual method should not depend on the specific classifier it explains.
\Cref{tab:ablation_classifier} therefore keeps CounterFlow unchanged and only substitutes $f$, comparing the RepVGG-A2 used throughout the paper with a deliberately mis-specified variant, namely the same network evaluated without its input normalization, which drops its accuracy to $86.3\%$, and with an adversarially robust PreActResNet-18 \citep{gowal2021improving}.
The mis-specified classifier is explained as well as the original one: validity stays at $100\%$ and both FID scores remain within two points of the default, indicating that CounterFlow does not require the explained model to be accurate, only that it defines an exploitable decision boundary.
The robust classifier behaves differently: validity drops to $83.5\%$, while FID$_{\mathrm{t}}$ reaches the lowest value we observe ($55.72$).
This is consistent with what adversarial training is designed to achieve, namely flattening the decision boundary and suppressing the small non-robust directions that a gradient-based guidance term can exploit, so that $\Lcal_{\mathrm{Cla}}$ pushes less strongly and a fraction of the samples never crosses the boundary.
The counterfactuals that do cross must therefore carry a genuine, large-scale semantic change, which is what brings them closer to the target distribution; note that the $16.5\%$ of samples that remain in the source class can only inflate FID$_{\mathrm{t}}$, so this value is, if anything, pessimistic.
Explaining a robust model is thus harder for CounterFlow, but yields more plausible counterfactuals, which is coherent with diffusion-based methods that explicitly require an adversarially robust classifier \citep{augustin2022diffusion}.

\begin{table}[H]
    \centering
    \caption{Effect of the explained classifier $f$ on CIFAR-10. Accuracy (Acc.) is the test accuracy in \% of the explained classifier itself, not a property of the generated counterfactuals.}
    \label{tab:ablation_classifier}
    \setlength{\tabcolsep}{4pt}
    \begin{tabular}{lcccc}
        \toprule
        Explained classifier $f$ & Acc. & \textbf{CV}$\uparrow$ & \textbf{FID$_{\mathrm{t}}$}$\downarrow$ & \textbf{FID$_{\mathrm{s}}$}$\downarrow$ \\
        \midrule
        RepVGG-A2 \emph{(default)} & 94.8 & 100.0 & 62.54 & 117.03  \\
        RepVGG-A2 w/o input normalization & 86.3 & 100.0 & 64.03 & 115.30 \\
        Robust PreActResNet-18 \citep{gowal2021improving}$^{\dagger}$ & 87.5 & 83.5 & 55.72 & 119.89 \\
        \bottomrule
    \end{tabular}
\end{table}

\subsection{Choice of datasets}
\label{sec:supp_datasets}

We restrict our experiments to MNIST and CIFAR-10, the largest settings in which all compared methods can be run with their official implementations, the same classifier, and the same evaluation protocol.
For instance, extending this comparison to ImageNet raises the following issues.

\textbf{Baselines.}
C3LT requires a separate model for each pair of classes, \ie, $499{,}500$ models for the $1{,}000$ classes of ImageNet.
Hence, its ImageNet experiments are qualitative, restricted to five pairs of classes, and use latent codes sampled from a pretrained GAN rather than real images \citep{khorram2022cycle-consistent}.
LeapFactual relies on a latent space in which inputs can be faithfully encoded, and each of its leaps requires decoding the latent code and querying the classifier; its only large-scale experiment is a binary task on images generated by a pretrained GAN \citep{cao2025leapfactual}.
Diffusion-based methods do apply to ImageNet, but at a high cost: on CIFAR-10, FastDiME already requires about $900\times$ more FLOPs per counterfactual than CounterFlow (\Cref{tab:results}), and ACE is evaluated on only three pairs of ImageNet classes \citep{jeanneret2023adversarial}.

\textbf{Evaluation.}
Our FID$_{\mathrm{s}}$ and FID$_{\mathrm{t}}$ are computed against the test images of a single class, whereas ImageNet provides only $50$ validation images per class, which makes such estimates unreliable.
Moreover, the number of ordered source-target pairs grows from $90$ to $999{,}000$.
As noted by \citet{jeanneret2023adversarial}, counterfactual explanation methods still lack an appropriate testbed on such complex datasets.

\textbf{CounterFlow.}
In principle, CounterFlow applies to ImageNet by adapting its U-Net architecture.
However, unlike the baselines, which build on generative models pretrained at a considerable computational cost, CounterFlow would have to be trained from scratch in pixel space, which exceeds our computational budget.
% Optionnel (point 4) :
Moreover, the number of source--target contexts $(y_0,y_1)$ grows quadratically with the number of classes, so that each context would be seen less often during training.
We leave this large-scale study to future work.

\clearpage
\subsection{Additional Results}
\label{sec:supp_additional_results}

\begin{figure*}[h!]
    \centering
    \includegraphics[width=0.85\linewidth]{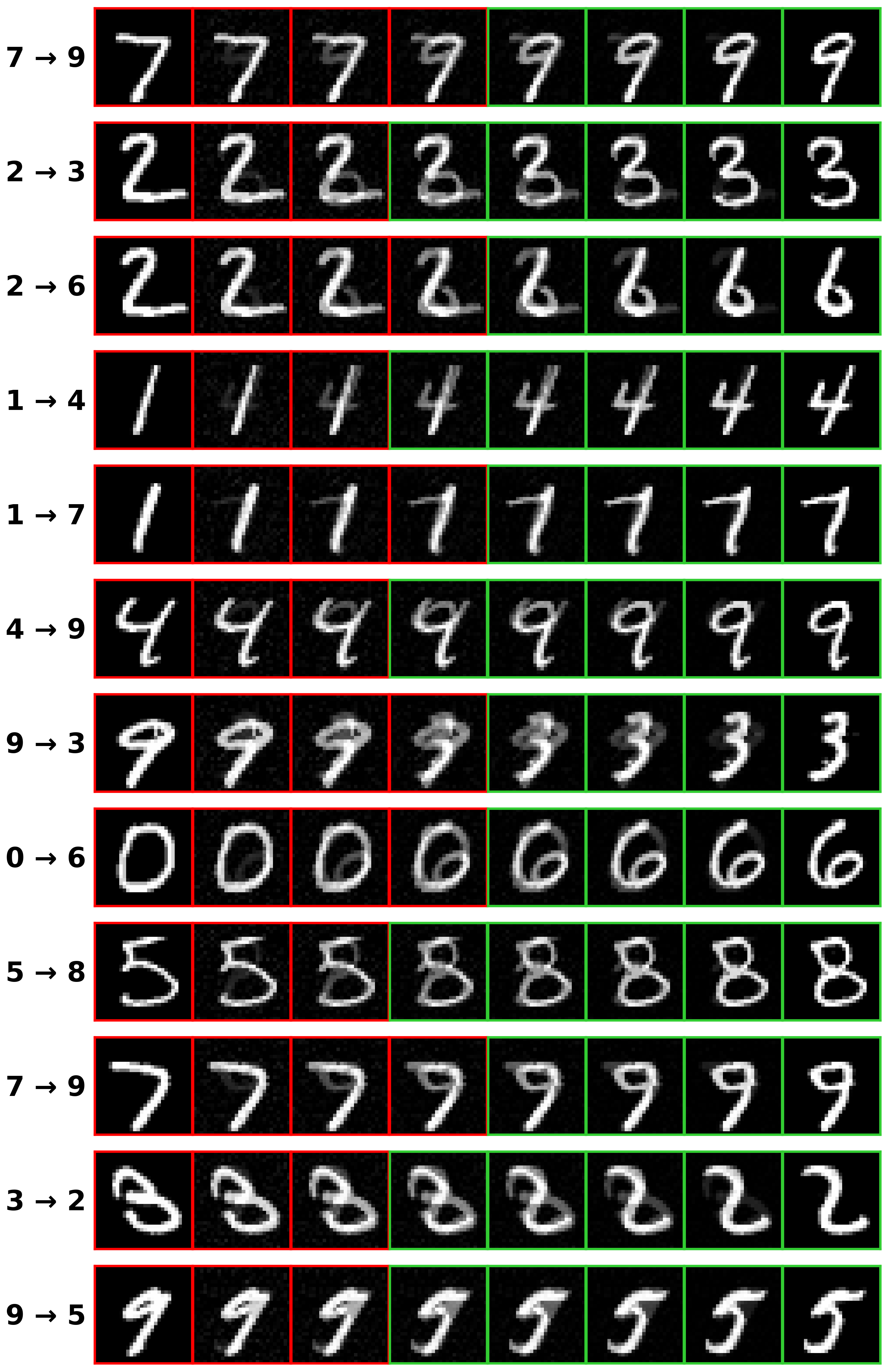}
    \caption{Additional CounterFlow trajectories on MNIST, from source to target class. Border color indicates the classifier's prediction along the path: \textcolor{red}{red} for the source class, \textcolor{green!50!black}{green} for the target class, and \textcolor{orange}{orange} for any other class.}
    \label{fig:additional_trajectories_mnist}
\end{figure*}

\begin{figure*}[ht]
    \centering
    \includegraphics[width=\linewidth]{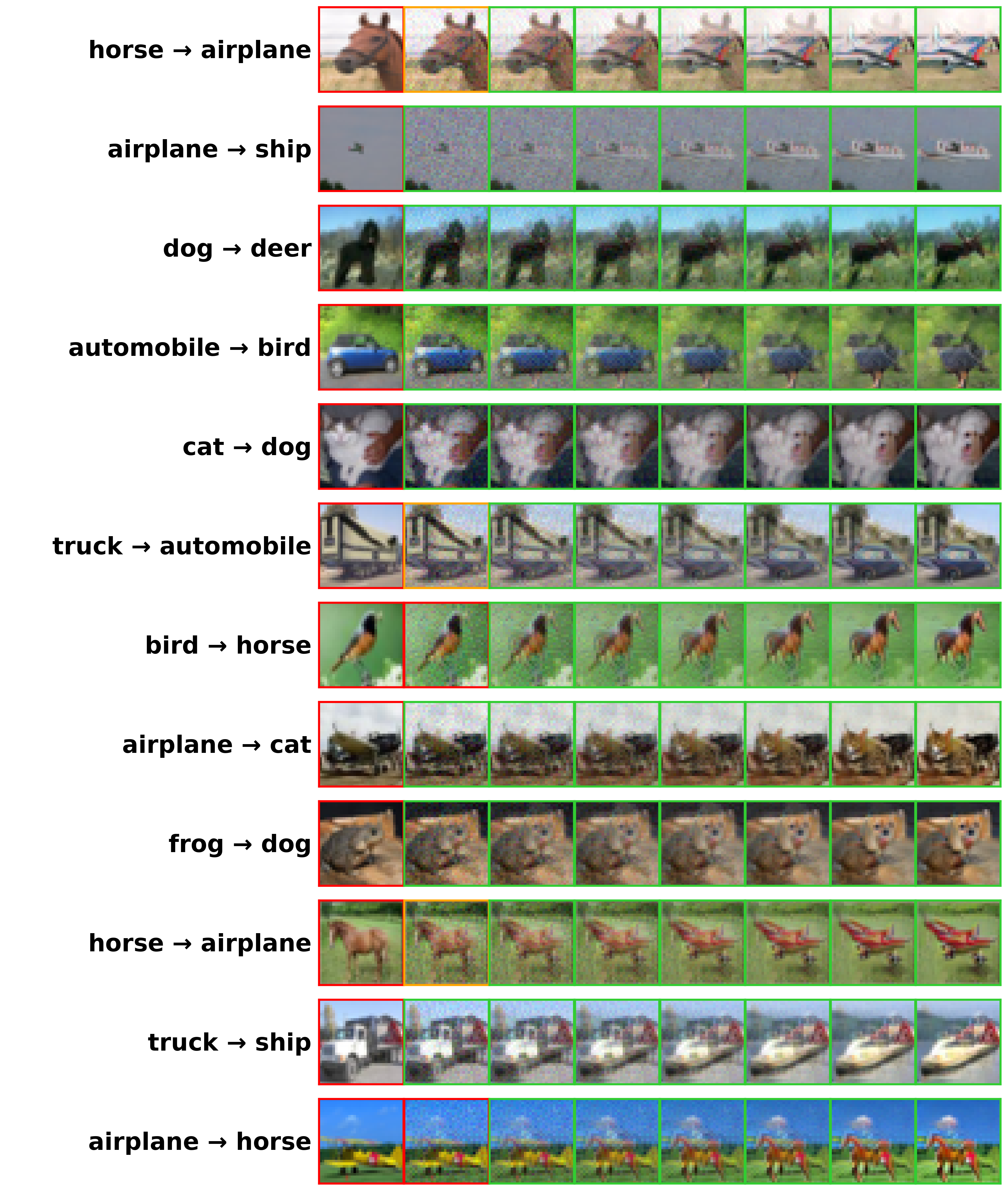}
    \caption{Additional CounterFlow trajectories on CIFAR10, from source to target class. Border color indicates the classifier's prediction along the path: \textcolor{red}{red} for the source class, \textcolor{green!50!black}{green} for the target class, and \textcolor{orange}{orange} for any other class.}
    \label{fig:additional_trajectories_cifar10}
\end{figure*}

\end{document}